# PPMF: A Patient-based Predictive Modeling Framework for Early ICU Mortality Prediction

**Mohammad Amin Morid, PhD Candidate [a]; Olivia R. Liu Sheng, PhD[a]; Samir Abdelrahman, PhD [b]**

[a] *Department of Operations and Information Systems, David Eccles School of Business, University of Utah, Salt Lake City, UT, USA*
[b] *Department of Biomedical Informatics, University of Utah, Salt Lake City, UT, USA*

## Abstract

*To date, developing a good model for early intensive care unit (ICU) mortality prediction is still challenging. This paper presents a patient based predictive modeling framework (PPMF) to improve the performance of ICU mortality prediction using data collected during the first 48 hours of ICU admission. PPMF consists of three main components verifying three related research hypotheses. The first component captures dynamic changes of patients' status in the ICU using their time series data (e.g., vital signs and laboratory tests). The second component is a local approximation algorithm that classifies patients based on their similarities. The third component is a Gradient Decent wrapper that updates feature weights according to the classification feedback. Experiments using data from MIMICIII show that PPMF significantly outperforms: (1) the severity score systems, namely SASP III, APACHIE IV, and MPM0III, (2) the aggregation based classifiers that utilize summarized time series, and (3) baseline feature selection methods.*

## Introduction

Accurate mortality prediction impacts medical therapy, triaging, end-of-life care, and many other aspects of ICU care [1]. Both intensive caregivers and patients' families can benefit from decision support for the crucial decision on aggressive or supportive treatment. Also, unexpected deaths which are still common despite evidence that patients often show signs of clinical deterioration hours in advance, can be detected [2].

Currently, most of the ICU mortality prediction tools that hospitals utilize are based on severity score systems [1,3], which have poor prediction accuracy on an individual basis. One of the main reasons that severity scores have limited ICU mortality prediction power is that their main intended use was for comparing groups of patients and to stratify patient populations among hospitals. Recently, classification methods that apply advanced machine learning techniques to time series ICU patient data have been proposed to successfully improve ICU mortality prediction over severity score systems.

Some of the most accurate extant ICU mortality predictive models utilize aggregation functions (e.g., average, worst and best values of *heart rate* in the first 48 hours after ICU admission) of time series data and all of the patients in training sets to estimate a new ICU patient's mortality outcome [4]. This study hypothesizes that the information omission due to such aggregation functions limits the accuracy of their ICU mortality prediction.

Medical experts as well as clinical decision support systems often rely on similar prior patient cases to explore hypotheses and make diagnosis for a new patient [5]. They often use case based reasoning to find similar patients they have visited before to treat the current patient [6]. This allows them to explore hypotheses about the current patient. Based on this idea, we developed a case-based reasoning framework for ICU mortality prediction using similarities amongst ICU patients' time series data. We hypothesize that the information richness endowed by time-series ICU patient data warrants the potential for improving the accuracy of mortality prediction based on more advanced time-series similarity measures and feature weight assignments.

Based on the aforementioned hypotheses, this paper proposes a patient-based predictive modeling framework (PPMF) that incorporates 1) an ICU patient time-series data standardization and representation scheme, 2) a similarity-based predictive model and 3) an error-reduction component for feature weight adjustment. To validate the effectiveness of this framework, we conducted empirical analysis to validate these hypotheses. The analyses have shown PPMF significant improvements compared to those of 1) severity score systems, 2) prediction methods used by the winner [7] of the PhysioNet/Computing in Cardiology Challenge 2012, and 3) baseline feature selection methods. Since PPMF makes effective prediction using only the first 48 hours patients' dynamic (i.e., time series) data, it may assist in early detection of patients who are in high risk of mortality as well as those who are expected to have stable or improving situation. Of note, the method proposed by the winner of the PhysioNet/Computing in Cardiology Challenge 2012 is referred as CCW (Cardiology Challenge Winner) in the rest of the paper.

**Related Work on ICU Mortality Prediction**

Most of the ICU mortality prediction researchers have focused on using illness severity index scoring systems to fulfill the task. The Acute Physiology And Chronic Health Evaluation (APACHE), the Simplified Acute Physiology Score (SAPS) and the Mortality Probability Model (MPM) are some of these scoring systems [8]. The APACHE score was designed based on the domain knowledge of an expert clinical panel to select the variables and assign severity score to each. Similar to that, the SAPS metric was designed by clinical experts aiming to match the APACHE performance using a simpler model. Using the same expert knowledge variables, the MPM metric uses forward stepwise selection methodology to select important variables and expert knowledge to for severity score.

As one of the first studies based on a data driven approach to improve ICU mortality prediction models and features, APACHE III utilizes multivariate logistic regression to appropriately assign weights to each model feature. Following it, APACHE IV uses step-wise feature selection techniques to select a subset of covariates in the model [9]. In spite of the improvements, the models lack sufficient prediction accuracy required at the patient level [3,8,10,11].

To enhance the performance of ICU mortality prediction more sophisticated machine learning methods have been utilized recently. The PhysioNet/Computing in Cardiology 2012 Challenge aimed to provide a benchmark environment for ICU mortality prediction. [10]. The competition was on two prediction tasks. This study focuses on the first task which was to predict mortality of ICU patients before hospital discharge as a binary classification problem. The winner [7] (i.e., CCW) proposed a new Bayesian ensemble scheme comprising 500 weak decision tree learners which randomly assigned an intercept and gradient to a randomly selected single feature. The parameters of the ensemble learner were determined using a custom Markov chain Monte Carlo sampler. The best score achieved by the CCW method for this task was 53.53%. This score is computed as the minimum between precision and recall on the mortality (positive) class. The performance of the severity score systems such as SAPS was only 31.25%. Other top teams applied different algorithms on the aggregation functions such as average, minimum and maximum extracted from input predictors to outperform the severity score systems [10]. After the competition researchers studied advanced machine learning methods on both task. Ryan et al. [12] trained a feed-forward neural network initialized by a deep Boltzmann machine and fine-tuned using an efficient approximation ensemble method. They improved the performance of the winner in task two, but they were not successful in beating the best performance of task one. To the best of our knowledge, the competition CCW method is the most successful ICU mortality prediction using patients' first 48 hours of data. Therefore, this study benchmarks the proposed framework against this method for evaluation.

**Method Background**

**Case based reasoning.** To solve a new problem, a case-based method can be developed to retrieve similar cases and adapt them to fit the new problem. A case based reasoning frameowrk[13] consists of four components: input (the problem), process (the similarity based case retrieval), output (the solution (e.g., diagnosis or cost)) and feedback (an evaluation of the solution). The most important component is the process that measures the new case relevance to the previously observed cases. This is done by calculating the similarity between the case at hand and those observed in the past. More similar the cases are to the focal case, more accurate the proposed solution is. The proposed framework utilizes a case based reasoning approach.

**Similarity based classification.** Similarity-based classifiers predict the class label of a test instance based on its similarity to a set of labeled training instances [14]. The pairwise similarity between instances can be computed using different methods. A typical similarity-based classification approach is to treat the given similarities as the reversed function of dissimilarities that is equal to instances' distance in some Euclidean space. This approach can be divided into two categories [15]. The first category directly uses instances in a multidimensional Euclidean space according to the given (dis)similarities. This approach considers each sample as a set of features, and the dissimilarity is computed according to pairwise comparison among features. K nearest neighbor (*k*-NN*)* is one the well-known techniques implementing this approach. The second category converts the similarities to be kernels and applies kernel classifiers on the data. To do so, kernel based algorithms, specially support vector machine (SVM), are applied on Euclidean space that is converted into positive semidefinite (PSD) kernels. This study uses the first approach to classify ICU patients for mortality.

**Feature weighting.** In healthcare problems features (e.g., diagnosis, symptoms and lab results) have different effects on patients based on the problem requirements. Researchers have investigated empirically several algorithms on the weight setting of *k*-NN algorithms. Some studies suggest that the weight of all features should be acquired by domain knowledge from experts. The main limitation of this approach is that manual weights are not available in all contexts as they are highly dependent on availability and knowledge of the domain experts. Supervised weighting is another approach that has been utilized in many methods which automatically assigns weights. There are two schools of methods for supervised feature weighting: wrapper and filter [16]. Wrappers assign weights to features based on the feedback they receive from the classifiers iteratively, while filters do not have such iterative feedback cycle. As an example of filter models, researchers [17] proposed an approach of assigning continuous weights to all attributes simply by their information gain values. The main weakness of such filters is that they treat each feature independently without considering their composite effects. Weak features if combined appropriately may become a strong predictor. The advantage of the wrappers is that they neither rely on domain knowledge nor ignore the power of feature combination. Weights are assigned to features while the classification (prediction) algorithm observes prediction errors.. Wrappers have outperformed filters according to past research [18], since the more error data is obtained and used, the more likely the weights can be adjusted to reduce prediction errors. This study uses a wrapper approach to update features' weights.

## Method: A Patient Based Predictive Modeling Framework for ICU Mortality Prediction (PPMF)

Using a patient time-series data standardization and representation scheme, a similarity-based predictive model and an error-reduction feature weight adjustment component, the proposed patient-based predictive modeling framework (PPMF) simulates the decision making process of physicians for clinical decision predictions or ICU mortality prediction. Figure 1 graphically depicts the data and control flow in an implementation of this framework. First, the input data is framed into time series format based on a specific time frame. These features are based on historical data of the patients, which contains multiple records for each patient (e.g., visits or ICU records). Second, a similarity based classifier is applied on the data to find similar patients to a focal patient. Third, prediction is made based voting or averaging of the similar patients to predict patients' outcome as in k nearest neighbor algorithm. Then, the performance of the algorithm is evaluated against the actual outcome and the error is computed. Finally, in the fifth step, features' weights are updated based on the computed error to reduce the classifier error. The four-step cycle in this framework repeats until error reductions cannot be found by the Gradient Descent search method. The contribution of this paper is embedded into the first, second and the last components of the proposed framework (bolded in Figure 1). Grounded on an underlying theoretical hypothesis, each component is created to validate the hypothesis. The rest of this section elaborates on these components and their underlying hypotheses.

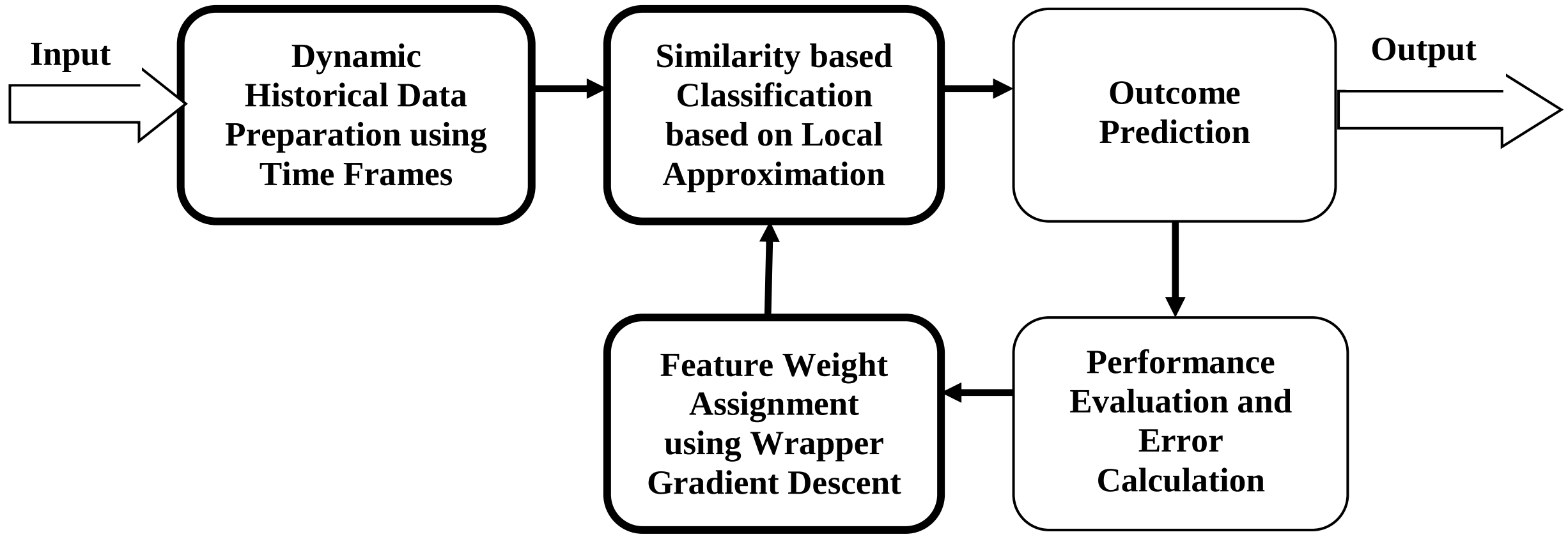

Figure 1: Abstract view of the proposed PPMF framework

**Dynamic Historical Data Preparation using Time Frames.** In general, there are two types of data for any healthcare analysis including ICU mortality prediction. Static data consists of patient's features like gender, race and even weight and height in many cases. Dynamic data consists of multiple records for each patient on such features as lab results, diagnosis codes, input and output of visits and ICU records.

Applying data mining methods on dynamic data is difficult, since each patient has multiple records (e.g., visits). Most data mining techniques require treating each patient as a single record consisting of a specified set of features, instead of multiple records for each feature. Therefore, the data should be organized in a way that meets the requirements of these techniques. To address this limitation, most studies look for a method to summarize dynamic features by some aggregation functions to give a static snap-shot based on the dynamic records. This is done by aggregation functions like minimum, maximum and average. Severity score systems and the machine learning based ICU mortality predictors use such measures. For example, they consider the worst value of blood pressure, glucose, sodium and respiratory rate based on their minimum or maximum. These snap-shot, aggregations cannot adequately capture the historical (time variant) behavior of patients. Two patients with very different situations might have the same aggregation values. Therefore, we believe that all historical (time series) data should be considered at a relatively fine-grain level to improve the performance of the current state of ICU mortality prediction methods.

For ICU mortality prediction, there is a huge number of records for each patient in different time points, where one patient may have records at time *t* but another patient does not. Therefore, there is a need for a unified time frame format to standardize the patients' time series data. One way to achieve this is to get the average of a feature's value per hour which results in 48 data points for each feature. However, the results showed that 48 hours' time framing gives an extremely sparse dataset (i.e., more than 60 percent sparsity). Therefore, we used the next best choice, which was getting the average of a feature's value per two hours to have 24 data points for each feature (28 percent sparsity). As a result, each of the 36 time-series features are compared for each two patients in terms of their 24 values. Similar decisions in time framing can be made for other healthcare challenges. For instance, for prediction of patients' diabetes two years ahead of actual diagnosis based on past three years visits' data, the time series framework can be recorded in weekly or monthly time frames depending on how often patients has been visited.

*Hypothesis 1: Using standardized historical patients' data, a predictive model will outperform the model with ICU mortality predictors that are based on static or aggregation functions*. To test this hypothesis, PPMF is compared with the same framework that uses the aggregation functions utilized by CCW [7]. Also, the contribution of static and dynamic features to the framework performance is assessed.

**Similarity based Classification based on Local Approximation.** Non similarity based methods give general approximation of the inputs (e.g., patients) behavior, while similarity based methods (e.g., nearest neighbor techniques) seeks local approximation based on similar inputs [15]. The latter type of techniques is usually used when the target function is complex and hard to achieve an effective general approximation. In the case of ICU mortality, it is hard to find general rules or patterns to model the mortality behavior [3]. Previous studies [11] argued that one of the main disadvantages of severity score systems is that they are based on general scoring patterns (rules), while ICU mortality problem needs patient specific techniques. Therefore, we believe that similarity methods can best fit complex health predictive problems such as ICU mortality prediction, as they do not rely on general scoring methods or approximation models. For local approximation in this study, feed forward weighted k nearest neighbors is utilized. More specifically, most patients similar to a focal patient are found based on a weighted similarity formula (elaborated in the next section). Then the label of the patient is predicted based on the most dominant label among her closest neighbors. The similarity formula is updated based on the classifier's error iteratively using a Gradient Descent search method.

*H2. A patient-similarity based method outperforms non similarity method for ICU mortality prediction*. To validate this hypothesis, the performance of the PPMF is compared with that of the illness severity score systems including SASP III, APACHIE IV, and MPM0III as well as with that of the Physionet/computing in cardiology 2012 challenge winner that utilized non similarity based machine learning method.

**Feature Weight Assignment using a Gradient Descent Wrapper.** To update features using supervised wrapper techniques there are a varieties of wrapper methods proposed in different studies. Most popular method is to increase learning speed by exploiting knowledge of the error function's gradient [19]. Feature weights are optimized using a Gradient Descent search algorithm to minimize total classification error on the training set. The derivative of this error with respect to each feature weight is used to guide the gradient procedure.

For this study, the feature weights are updated based on the classification feedback of the proposed similarity based classifier (i.e., wrapper approach). Similarity based classification such as *k*-NN is named lazy learner. This learner does not have a training model and when a test instance (i.e., patient) arrives, the learner compares it to all instances in the training dataset to find the nearest neighbors and estimate its label. In PPMF framework, feature weights are updated using Gradient Descent wrapper [20,21]. During the training procedure, the Gradient Descent wrapper uses the derivative of output error and a learning rate that determines the effect of a single misclassified instance. This error is defined as the squared sum over all training instances and all class labels (i.e., dead or alive until hospital discharge) of the difference between the desired output and the computed output. Therefore, the derivative of the error for each instance is computed with respected to the feature weights.

*Hypothesis 3: The Gradient Descent wrapper outperforms methods that consider the same weights for all features as well as baseline approaches for feature weighting.* To show the importance of the feature weighting approach in the proposed framework, we compare its performance against the same framework that uses no feature weights, manual feature weights and supervised feature weighting methods (i.e., filters) including chi square, information gain and gini index.

## Evaluation

**Data.** The Multi-parameter Intelligent Monitoring for Intensive Care (MIMIC) [22] database was created for facilitating development and evaluation of ICU decision-support systems. The latest version of this dataset is MIMIC III, which contains over 58,000 hospital admissions for 38,645 adults and 7,875 neonates. The data collects patent information from June 2001 - October 2012.

In this study, the performance of PPMF is compared with the PhysioNet/Computing in Cardiology Challenge 2012 competition winner (i.e., CCW) [7]. Therefore, we used the same experimental setup in the competition by filtering patients to 22,561 patients who are younger than 16 years old and remained in the ICU for 48 hours or longer. MIMIC III is a more complete dataset with a larger number of patients with reduced error and incompatibility comparing to MIMIC II, which was the original dataset used in the competition.

The data input consists of time series data of 36 variables (i.e., dynamic features) extracted from the first 48 hours of patients' ICU stay, shown in Table 1, plus four static features (i.e., age, gender, height, and initial weight). The target variable is a binary feature showing that whether or not the patient eventually dies in the hospital before discharge. As illustrated before, 24 predictors were extracted from each dynamic variable, where each predictor shows the average value of the dynamic variable in a specific 2-hour time window.

Table 2 shows the data distribution over the target variable in the cleansed data. As seen, while almost half of the ICU patients have died eventually, most of the deaths happened out of hospital. The problem analyzed in this study is the prediction of in hospital mortality, which has 18% of the positive class against 82% of the negative class.

To ensure meaningful performance comparisons based on the same experimental setting as PhysioNet/Computing in Cardiology Challenge 2012, this study does not use all enriched information of patients in the MIMIC III dataset. The dataset has a variety of other information about patients such as diagnosis codes, reason of admission, microbiology information and spectrum of lab results that could be of value in future research.

**Experimental Setup.** For estimating and tuning the parameters including the best k, the best learning rate and filter approach feature weights, 50% of the data was used as the development set, while the rest remained unseen for validation set. In all development set experiments, k = 10 consistently led to good prediction performance and hence was chosen for finding similar patients to a focal patient. Also, 0.3 was chosen for the learning rate after trying different values. In experiment 1, SAPS III, APACHIE IV and MPM III models are those introduced in previous studies [23, 9, 24]. Also, CCW was re-implemented and applied to the cleansed data. In experiment 2, the PPMF was applied on the six aggregation functions used in CCW method including minimum, maximum, median, first, last and number of values as another baseline. Manual feature weights in the last experiment are the same weights that are used for calculation of APACHIE III score. Also calculation details for information gain, gini index and chi square can be found in the literature [25].

Table 1: Input predication features.

| Dynamic time series variables | | | Static variables |
|---|---|---|---|
| Invasive (diastolic) | Cholesterol | Na (Serum sodium) | Age |
| Invasive (mean) | Creatinine | PaCO2 | Gender |
| Invasive (systolic) | FiO2 (Fractional inspired oxygen) | PaO2 | Weight |
| Non-invasive (diastolic) | Glasgow Coma Score (GCS) | pH | Height |
| Non-invasive (mean) | Glucose | Platelets | |
| Non-invasive (systolic) | HCO3 (Serum bicarbonate) | Respiration rate | |
| Albumin | HCT (Hematocrit) | SaO2 | |
| ALP (Alkaline phosphatase) | Heart rate | Temperature | |
| ALT (Alkaline transaminase) | K (Serum potassium) | Troponin-I | |
| AST (Aspartate transaminase) | Lactate | Troponin-T | |
| Bilirubin | Mg (Serum magnesium) | Urine output | |
| BUN (Blood urea nitrogren) | Mechanical ventilation | WBC (White blood cell count) | |

Table 2: Data distribution over the target variable (mortality).

| | Num of Patients | Avg Hospital Stay | Avg ICU Stay |
|---|---|---|---|
| **Total** | 22561 | 11.78 | 5.49 |
| **Died** | 10584 (47%) | 13.87 | 5.53 |
| **Died out of Hospital** | 6516 (29%) | 13.39 | 4.86 |
| **Died in Hospital** | 4068 (18%) | 13.50 | 6.60 |
| **Died in ICU** | 3240 (14%) | 10.53 | 7.49 |

**Evaluation Setup.** In all experiments, 20-fold cross validation, over the 50% unseen data, was used to evaluate the performance of each method. Classification performance was measured according to the average precision, recall, and F-measure across the 20 folds, which are the most popular classification performance measures.

F-measure is the measure of choice for hypotheses testing. For statistical significance test of all experiments, first we applied the Friedman's test to verify differences among multiple classifiers. Comparisons among the classifiers according to Wilcoxon Signed-Rank test were made to determine if the Friedman's test was significant at an alpha of 0.05. This statistical evaluation approach is aligned with the method recommended by Demsar [26].

**Results**

**Experiment 1: Similarity based Classification based on Local Approximation.** Figure 2 shows that the F-measure for the proposed ICU mortality classifier performed significantly better than the CCW method and all baseline severity score systems including SASP III, APACHIE IV, MPM0 III (F-measure = 66% versus 55%, 33%, 32% and 30% respectively; $p<0.001$ for all comparisons).

**Experiment 2: Dynamic Historical Data Preparation using Time Frames.** Figure 3 shows that the F-measure of the proposed PPMF ICU mortality classifier performed significantly better than the same classifier applied on CCW aggregation functions (F-measure = 66% versus 60%; $p<0.001$). Also, the proposed classifier has lower F-measure of 61% and 8% if we remove static and dynamic features respectively ($p<0.001$).

**Experiment 3: Feature Weight Assignment using the Gradient Descent Wrapper.** Figure 4 shows that the F-measure for the proposed ICU mortality classifier performed significantly better than the classifier using different feature weighting methods including chi square, information gain and gini index, manual weighting and no feature weights. (F-measure = 66% versus 40%, 48%, 42%, 35%, 28% respectively; $p<0.001$ for all comparisons).

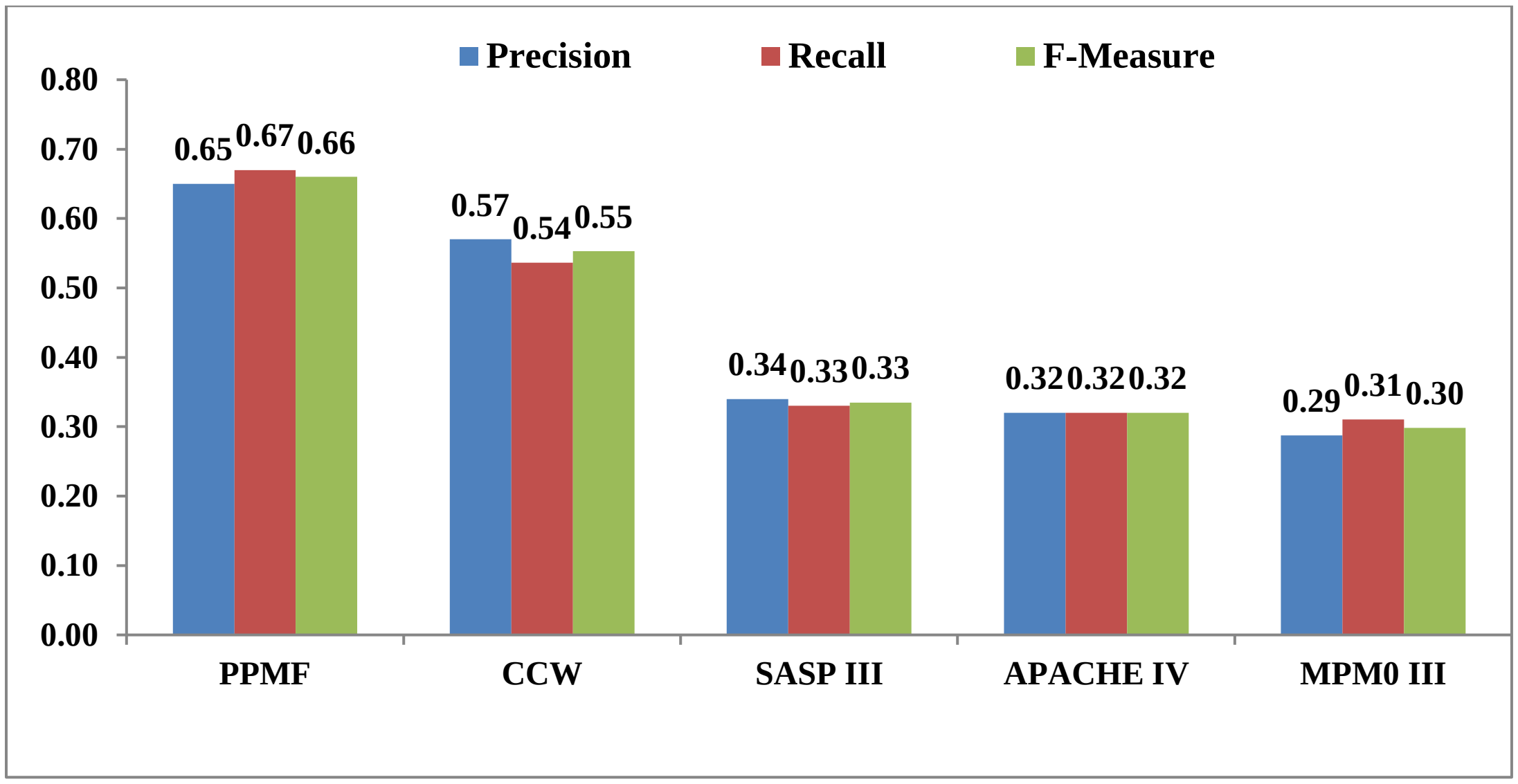

Figure 2: Average precision, recall and F-measure of PPMF comparing to other non-similarity based classifiers.

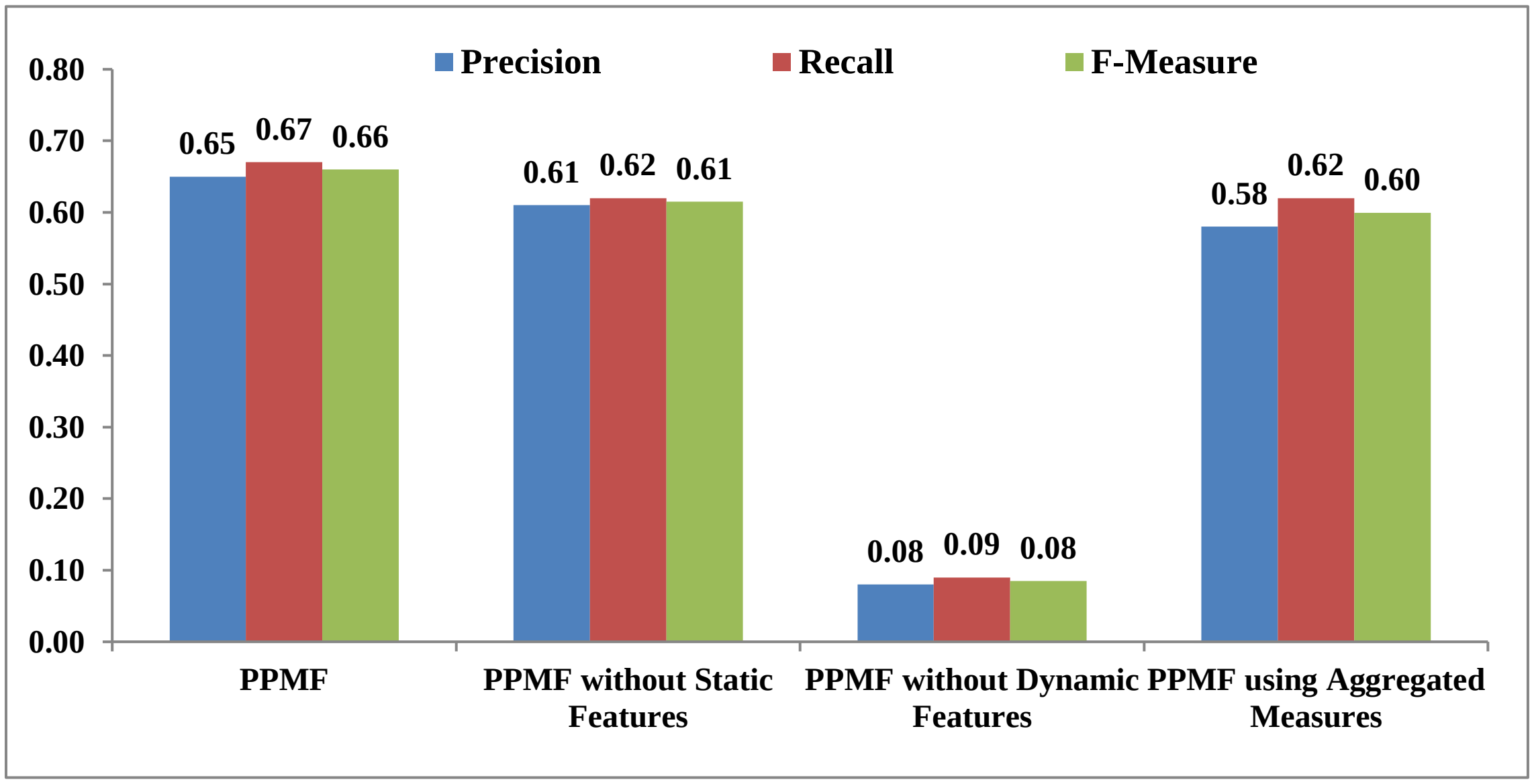

Figure 3: Average precision, recall and F-measure of PPMF comparing to the classifier with and without static and dynamic features.

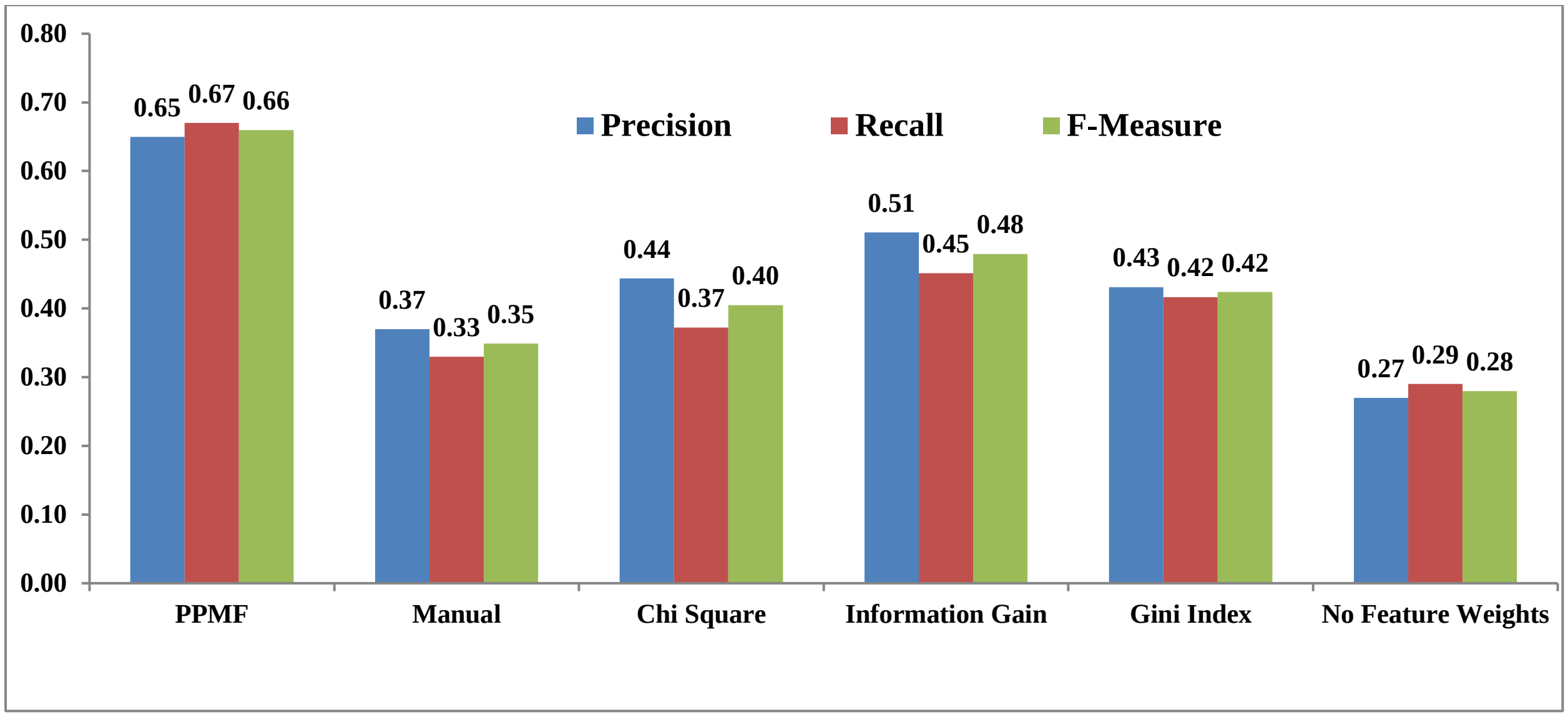

Figure 4: Average precision, recall and F-measure of PPMF comparing to other non-similarity based classifiers.

## Discussion

We conducted three experiments to test different hypotheses. The first experiment showed that similarity based approach outperformed extant machine learning methods as well as the severity score systems. The reason for the poor performance of severity score systems is that they are not designed to give patient specific prediction; instead, they are more appropriate for population based recommendation especially on a daily basis. These systems are adequate predictive models. Moreover, the reason that the baseline CCW machine learning method was outperformed by PPMF is that CCW is designed to give general approximation on the data. Accurate ICU mortality prediction needs to consider unique behavior of each patient and monitoring her status during consecutive time periods. PPMF uses local approximation models that afford the proposed classifier flexibility to tune according to each patient and her local patients with similar behavior.

The second experiment showed the importance of using patient's time series data rather than aggregation function to appropriately capture dynamic behavior of patients' status. These functions have less accuracy, since summarizing the data comes with information loss. Consequently, the six aggregation functions can be similar for totally different patients. On the other hand, PPMF captures dynamic changes of patients' status and their diversity using the proposed 24 data points' framework. Moreover, this experiment shows that both dynamic and static features are providing enough contribution, since PPMF outperforms the classifier without dynamic and static feature. Dynamic features can benefit from the static feature to have a higher accuracy.

The final experiment shows the significant contribution of the feature weighting, especially the weighting method used in this study. Not weighting features gives extremely poor performance, since all features are treated in the same way. In high dimensional problems such as ICU mortality prediction which involves 40 predictors, most of them having 24 data points, it is highly expected to see this happen. Manual feature weighting was not effective on this problem, since they are more suited for the problems with small numbers of dimensions and accurate domain knowledge. Moreover, weighting features based on statistical analysis (filter approach) showed poor performance, since they treat each feature individually without considering their combination. In a complex healthcare predictive analytics problem such ICU mortality prediction, there is a high chance of getting strong predictors by combining the weak ones. The PPMF's iterative update of feature weights using Gradient Decent (wrapper approach) had promising accuracy. This approach is highly beneficial when there are enough data instances to tune the weights. Although the wrapper approach has longer running time compared to the filter approach, its significantly better performance warrants the potential application of the PPMF for ICU mortality and other clinical decision predictions.

**Limitations.** The main limitation of this study was the use of single dataset for running and evaluating the proposed method. To validate whether the advantage of the local approximation method is specific to this data or can be generalized to other datasets, further experiments on other datasets can be beneficial. Also, to align with PhysioNet Cardiology competition the input time period was the first 48 hours data of ICU patients, and the output was their mortality before hospital discharge. Evaluating the method's performance on different input and output time periods can shed light on the method's performance in other ICU mortality prediction tasks.

## Conclusion

This paper proposes a patient based predictive modeling framework for ICU mortality prediction. The proposed method outperformed severity score systems including APACHE IV, MPM III and SASP III and PhysioNet Cardiology competition winner. The time series data standardization approach utilized in the framework outperformed aggregation methods that are commonly used for summarizing time series data. Gradient descent method used for assigning weight to features proved to be superior to other types of feature weighting methods that are not based on the classifier's feedback.

As a future direction, this framework can be applied on many healthcare prediction problems where the data is complex for general approximation models to fit (i.e., the data input consists of multiple records for each patient). Many methodological extensions to the proposed PPMF including enhancing time-series data standardization, and integrating distance and kernel approaches for local approximation could be fruitful future research directions as well.